# Interpretable Multiple Treatment Revenue Uplift Modeling

**Robin M. Gubela**
Humboldt-Universität zu Berlin
Spandauer Str. 1, 10178 Berlin
robin.gubela@hu-berlin.de

**Stefan Lessmann**
Humboldt-Universität zu Berlin
Spandauer Str. 1, 10178 Berlin
stefan.lessmann@hu-berlin.de

## Abstract

Big data and business analytics are critical drivers of business and societal transformations. Uplift models support a firm's decision-making by predicting the change of a customer's behavior due to a treatment. Prior work examines models for single treatments and binary customer responses. The paper extends corresponding approaches by developing uplift models for multiple treatments and continuous outcomes. This facilitates selecting an optimal treatment from a set of alternatives and estimating treatment effects in the form of business outcomes of continuous scale. Another contribution emerges from an evaluation of an uplift model's interpretability, whereas prior studies focus almost exclusively on predictive performance. To achieve these goals, the paper develops revenue uplift models for multiple treatments based on a recently introduced algorithm for causal machine learning, the causal forest. Empirical experimentation using two real-world marketing data sets demonstrates the advantages of the proposed modeling approach over benchmarks and standard marketing practices.

## Introduction

Big data and business analytics have gained significant interest in information systems and management research (e.g., Mikalef et al. 2020). Using related methods facilitates creating business value across a variety of applications (e.g., Chen et al. 2012) and advances business and societal transformations (e.g., Pappas et al. 2018). Predictive models guide decision-making by estimating a future event's occurrence based on large data amounts. In digital marketing campaigns, analysts use uplift models for customer targeting selection. In contrast to response models, uplift models consider the causal association between a treatment (e.g., a marketing stimulus) and a desired outcome (e.g., Gubela et al. 2019). To this end, an uplift model targets customers who are likely performing the incentivized task due to the reception of the treatment and for no alternative reason (e.g., Devriendt et al. 2018). Uplift models forecast the sign and strength of individual-level treatment effects (ITE) based on a unit's affiliation to a treatment or control group and corresponding covariates. Customer targeting decision-making follows the model's estimations. That is, analysts target customers according to their relative magnitude to respond favorably to a treatment.

Much progress has been made in the development of uplift models that address the *single treatment case* (e.g., Devriendt et al. 2018). That is, researchers examine which customers should obtain the treatment of interest. The single treatment case assumes the prevalence of one treatment only, which does not differ in its configurations. This could be a coupon with a fixed discount value. Furthermore, researchers often predict binary outcome variables, which is referred to as *conversion uplift modeling* (Gubela et al. 2019). Using related estimators, practitioners increase financial returns from targeting that correspond to dichotomous business outcomes (e.g., product purchases or lead generations). Despite the higher practical value compared to dichotomous responses, only a handful of studies examine continuous outcomes such as revenues, which the literature refers to as *revenue uplift modeling* (e.g., Gubela et al. 2020). As the execution of targeting campaigns requires financial resources, practitioners aim to increase returns on (marketing) investments. Therefore, the prediction and evaluation of continuous outcomes might serve as a more promising strategy.

In contrast to the single treatment case, the *multiple treatment case* refers to corporate practices where more than a single marketing incentive co-exist so that analysts allocate distinct treatments to different customers (e.g., Rzepakowski and Jaroszewicz 2012). Companies drive digital marketing strategies with multiple treatments to increase financial returns. In a couponing application, the multiple treatment case comprises coupon types with different discount values. A recent study consolidates the few available methods of multiple treatment uplift modeling, which yet estimate binary outcomes (Olaya et al. 2020).

A general concern of machine learning research is the trade-off between a model's predictive accuracy and interpretability. Clarifying the latter is challenging if the model processes huge and high-dimensional data sets as prevalent in online marketing. Often, non-linear functions (e.g., boosting algorithms) achieve higher accuracy than linear functions (e.g., least-square regressions), but are limited in their interpretability of model outputs (e.g., James et al. 2013). Uplift researchers typically aim to increase an uplift-specific measure of accuracy at the expense of interpretability. Thus, they extent non-linear machine learning algorithms toward forecasting the change in a unit's behavior (e.g., Rzepakowski and Jaroszewicz 2012). The prevalence of treatment and control groups further complicates the interpretability of model outputs. Few studies clarify an uplift model's estimation and evaluation and translate findings into tangible and understandable business metrics for decision support (e.g., Gubela et al. 2017).

The study contributes to the growing research of uplift modeling as follows. It extends conversion uplift modeling for single treatments by developing revenue uplift models for multiple treatments. Estimating and evaluating continuous outcomes better reflects campaign management practices compared to binary outcomes.

Given the prevalence of co-occurring treatments, the choice of an optimal treatment from an assortment of options calls for scientific analysis. The knowledge of a treatment's relative effectiveness is crucial as treatments usually differ in their persuasive impact and monetary value. To our best knowledge, our study pioneers in the field of multiple treatment revenue uplift modeling. Additionally, the paper is the first to apply the causal forest algorithm (Athey et al. 2019) to the multiple treatment case. To facilitate the interpretability of results, the paper conducts analyses of the model's estimations in terms of ITE distributions and variable importance per treatment. Also, the paper reveals details about the model's performance evaluation according to critical business metrics. It further empirically confirms the advantage of multiple treatment revenue uplift modeling compared to multiple treatment conversion uplift modeling and identifies the most effective treatments to inform decision-making. The analyses are based on two real-world marketing data sets, that is, a proprietary data set from a European online bookseller with six coupon types and a disclosed data set about US-based e-mail merchandising promotions for men and women.

## Background

The conceptual idea of uplift modeling relates to four types of customers, which we summarize as follows. *Untreated non-purchasers* have not been inferred with a treatment and have not bought a product as a result of their customer journey. *Treated non-purchasers* have received a treatment, but the treatment did not have a sufficiently large effect on their purchasing behavior to buy a product. *Untreated purchasers* have bought a product without interference, as they decided to buy it upfront or as part of their customer journey. *Treated purchasers* have purchased a product as a reaction toward the reception of a treatment.

Uplift models typically estimate purchase likelihoods of customers who were exposed to a treatment (i.e., a *treatment group*) and customers who were not exposed to a treatment (i.e., the *control group*) conditional on specific characteristics. Then, ITE are calculated as the difference between the conditional probabilities. Under the potential outcome framework of causal inference, randomized treatment allocation is a critical property as it complies with the conditional independence and overlap assumptions, which together are denoted as *strong ignorability* (Rosenbaum and Rubin 1983). While the former secures that the potential responses do not interdependent with the treatment allocation process conditional on features, the latter specifies that treatment reception probabilities are positive and below a value of "1" so that every unit might obtain a treatment (e.g., Rosenbaum and Rubin 1983). The stable unit treatment value assumption (SUTVA), known as the third assumption of the potential outcome framework, clarifies that a unit's outcome is agnostic to treatment allocation decisions of other units (Rubin 1980). The three assumptions expand to the multiple treatment case as shown by Lopez and Gutman (2017). Uplift models that comply with these assumptions facilitate unbiased estimates of treatment effects so that a unit's behavioral change is caused by a treatment.

## Related Literature

Many uplift researchers study the effectiveness of a single treatment on a binary outcome. The research objective in the single treatment case refers to examining which customers to provide a single type of offer. The binary outcome setting comprises the task to predict a 0/1-scaled response variable. Therefore, both the treatment and the outcome variables have two levels. We define this setting as *single treatment conversion uplift modeling* or *"ST-Conv"*. Examples of estimators in this field refer to causal trees and forests (e.g., Athey et al. 2019). Benchmarking studies consolidate estimators of ITE and compare their performances across data sets (e.g., Kane et al. 2014; Knaus et al. 2018).

Besides dichotomous responses, the single treatment case comprises predictions of continuous outcomes that provide more granular insights into a customer's level of commercial engagement. Practitioners prioritize continuous returns from financial investments. Therefore, estimating customer revenues follows a more useful target than click occurrences or product purchases. As in the ST-Conv setting, the treatment variable has two levels. While the outcome variable has more than two levels, we define the setting as *single treatment revenue uplift modeling* or *"ST-Rev"*. The few related studies propose different modeling strategies in the field of continuous outcomes (e.g., Gubela et al. 2020; Rudaś and Jaroszewicz 2018).

In contrast to single treatments, the multiple treatment case regards the existence of several marketing incentives (e.g., coupon types or paid search advertisements). It informs targeting decision-making based on the most effective treatment(s). Recent literature extends the idea of conversion uplift modeling to the space of multiple treatments. In contrast to the outcome variable, the treatment variable has more than two levels. Therefore, we define this setting as *multiple treatment conversion uplift modeling* or *"MT-Conv"*. Related methods are adapted decision trees (Rzepakowski and Jaroszewicz 2012), context-variant random forests (Zhao et al. 2017), meta-approaches (Zhao and Harinen 2019), logistic regression models (Lo and Pachamanova 2015), and reinforcement learners (e.g., Li et al. 2019). A recent benchmark study systematizes the methods toward the MT-Conv setting and provides an analysis of model performance across several empirical data sets (Olaya et al. 2020).

## Multiple Treatment Revenue Uplift Modeling

We associate revenue uplift modeling with the multiple treatment case to exploit the practical benefits of predicting continuous responses while considering the occurrence of several incentives. As both the treatment and the outcome variables have more than two levels, we define this setting as *multiple treatment revenue uplift modeling* or *"MT-Rev"*. Figure 1 presents the paper's conceptual framework, which considers the single and multiple treatment cases in conjunction with conversion and revenue uplift modeling, and emphasizes its focus on the novel setting. In the following, we extend the multiple treatment approaches of conversion uplift modeling toward revenue uplift modeling.

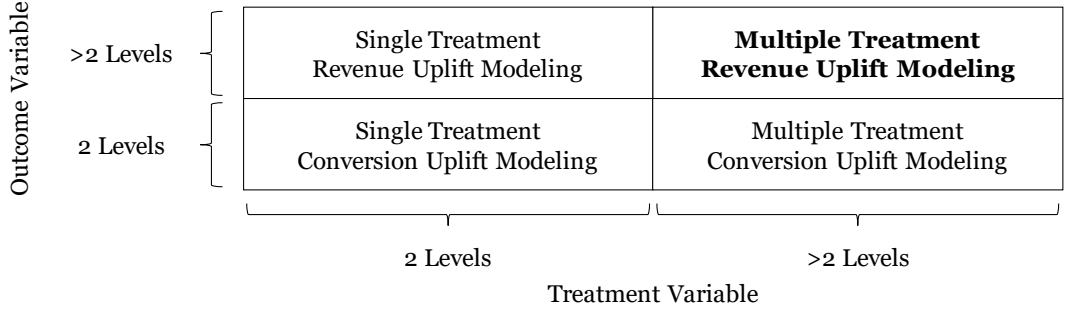

**Figure 1. Uplift Treatment/Outcome Matrix**

MT-Conv covers two modeling approaches to support the decision-making of customer/treatment allocation. The first approach combines the available treatments and compares the set of treatments against the control. Analysts alter the treatment variable so that its values are "1" or "0" for cases that obtained a treatment or no treatment, and use classification algorithms to estimate binary responses (e.g., Kane et al. 2014). An extension to MT-Rev is straightforward. In addition to the treatment variable manipulation, analysts estimate continuous outcomes. The multiple treatment problem resembles the single treatment setting as the data manipulation implies that distinctions between treatments are no longer possible.

The second approach does not alter the treatment variable. Instead, analysts compare each treatment against the control to determine the treatments that are more effective than the control in terms of specific performance metrics (e.g., response rates). Subsequently, analysts identify the best treatment(s) based on a comparison of a treatment's effectiveness relative to its alternatives (e.g., Lo and Pachamanova 2015). As the combined treatment approach, the treatment comparison approach generalizes to revenue uplift modeling in that it facilitates continuous response predictions. Let $t \in T = \{0, \ldots, K\} \in \mathbb{Z}_0^+$ denote a treatment $t$ from a set of $K$ treatments (with $t = 0$ as the control), and let $\Delta Y_r$ denote the difference in continuous outcomes. Figure 2 illustrates both revenue uplift modeling approaches for multiple treatments.

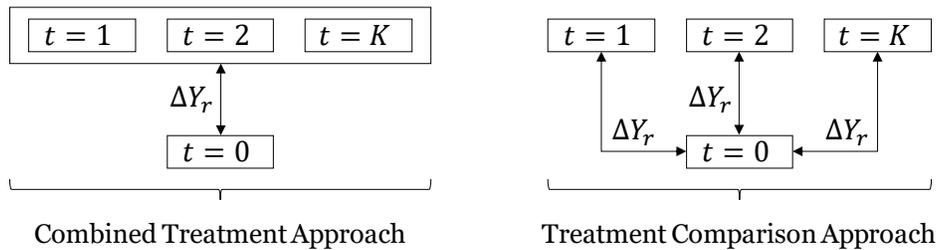

**Figure 2. Revenue Uplift Modeling Approaches for Multiple Treatments**

Both approaches allow an analyst to employ indirect or direct revenue uplift models. While direct revenue uplift models derive ITE as model output, indirect revenue uplift models require an analyst to specify separate models (i.e., a model per treatment and a model for the control) and subtract the conditional expectations of the control samples from the respective treatment samples. In a conversion setting, this is denoted the separate model approach (e.g., Lo and Pachamanova 2015), which can be adapted toward continuous outcome forecasts. An example of a related direct uplift model refers to the interaction term approach for multiple treatments, which weights a model's covariate space in favor of interaction terms between the covariates and treatments (Olaya et al. 2020) and predicts responses of continuous scale.

The combined treatment and treatment comparison approaches for revenue uplift modeling share some advantages. They benefit from methodological simplicity and facilitate the application of single treatment uplift models. Also, they allow the use of different regression algorithms due to their algorithm-agnostic design. However, they differ in that the combined treatment approach is less costly as it assesses whether all treatments are more effective than the control. The approach disregards (possibly large) differences in the effectiveness of treatments although some treatments might have negative effects on a subpopulation, which analysts want to avoid targeting. In contrast, the treatment comparison approach distinguishes between treatments but is more expensive as it requires a larger number of revenue uplift models.

We follow the treatment comparison approach and use the causal forest algorithm (Athey et al. 2019) with 1,500 trees for our empirical analysis, which can be regarded as a direct uplift model as it derives ITE. The causal forest algorithm is based on elements of random forests (Breiman 2001) such as bootstrap or subsample aggregation and greedy recursive partitioning but differs in the design of its splitting rule. Instead of favoring splits to minimize (mean-squared) errors of observed outcomes, as usually done is random forests, a causal forest splits the data to increase the heterogeneity of in-sample estimates. Therefore, the algorithm calculates gradients of parent node parameters to produce pseudo-outcomes that it pipelines to a regression task. It applies a forest-based weighting function that emphasizes units that are stored in the same tree leaves as a covariate's target value and uses the weighted outcomes to derive ITE (Athey et al. 2019). The causal forest algorithm uses local centering to liberate the influence of covariates.

We justify the choice of the causal forest as follows. First, the causal forest with local centering is a recognized ITE estimator that achieves competitive model performance (e.g., Gubela et al. 2020; Knaus et al. 2018). Second, the causal forest exhibits important statistical properties such as consistency and asymptotic normality (see Athey et al. 2019 for details). Third, it generalizes to estimate both dichotomous and continuous responses. Fourth, the causal forest clarifies how it arrives at its conclusions based on a variable importance analysis, for which it calculates an importance score per variable according to its relative number of splits at each node. To our best knowledge, the causal forest has yet only been applied in the single treatment case. While we mainly use the causal forest for the MT-Rev setting, we further add a causal forest-based MT-Conv benchmark to the empirical analysis.

## Empirical Analysis

### Data and Pre-Processing

We detail the data sets and pre-processing steps as follows. We received the first data set from an industry partner that consults European online shops with heterogeneous product groups. The data set is based on an e-couponing campaign of a European online bookseller. The campaign randomly allocates a coupon from a pool of six coupon types and a control (i.e., no coupon) to a customer session. The coupon pops up during the customer's online journey after either the third, sixth, or ninth pageview at the seller's online shop. A prospect/customer is supposed to receive a randomized treatment (control) with a 75% (25%) randomized probability that is fixed across sessions. Whereas a randomized subset of 15% of customer sessions is targeted, the remaining 85% of customer sessions are subject to a targeting process that uses a predictive model for customer selection. Our partner separates the data from the two processes into two folds for analysis. We are equipped with the former fold that contains the randomized treatment assignment data.

The treatments are coupon types with an absolute or percentage discount value, which we abbreviate with "A" and "P". The absolute discount values are 10€ and 15€, whereas the percentage discounts have values of 10%, 12%, 13%, or 15% relative to the final checkout amount. The coupons contain promotion codes that need to be inserted into a field during the checkout process. Coupons cannot be used to lower the prices of book purchases due to the fixed book price regulations, which limits the shop's flexibility to specify book prices. Thus, the coupons apply to alternative products such as music, videos, stationery, and home decor.

The coupon data set comprises 71,656 observations and 155 variables. It contains meta-variables that identify which treatment a customer received (if any), product purchases, and spend volumes. Apart from this, dozens of covariates measure a customer's interactions with the online shop. These include current session data (e.g., which page type the customer visited before, the session's daytime/nighttime and whether the shop visit is close to Christmas) and past session data (e.g., product purchases in the preceding week/month/year and session counts in the prior week). Moreover, device-related data (e.g., operating systems and mobile access) and location-based data is included. Further, the data set lists additional versions of these variables that were log-transformed by our industry partner.

Next to the coupon data, we use data from US-based e-mail merchandising promotions (Hillstrom 2008). The data has two advantages. First, it has open access, which facilitates reproducibility. Second, the data is based on a real-world use case and the variables are not pre-transformed or anonymized. Hence, the practical meaningfulness of our interpretations is not constrained. The data set has already gained attention in uplift research about single treatments (e.g., Kane et al. 2014). Two studies consider the data set for the multiple treatment case (Lo and Pachamanova 2015; Olaya et al. 2020). While the two treatments refer to promotional e-mail campaigns for men ("Men") and women ("Women"), respectively, the control indicates that a customer has not received an e-mail. The treatment assignment is based on a randomization process. That is, a randomized third of customers received the men mailing, the women mailing, and no incentive.

The data set consists of 64,000 observations of individual customers. In addition to the treatment group variable, it records meta-information about a customer's observed behavior in terms of website visits, product purchases, and his/her spending within a subsequent two-week period. Further covariates capture a customer's interactions with the website. They include the number of months since the previous product purchase, the amount of spending in the previous year, location-based information based on zip codes (i.e., urban, suburban, rural areas), whether customers converted in the previous year and purchasing channel information (i.e., phone, web, or multiple channel usage).

We pre-process the data sets as follows. For the coupon data, we drop 21 records with a coupon value of zero that seem to be erroneous. To facilitate the interpretation of results, we remove the log-engineered features and keep their original versions. We drop three variables with solely zero values. The pre-processed data set has 71,635 records and 69 variables, which do not have missing values. The number of observations varies per treatment. More specifically, the 10€ coupon, the 15€ coupon, the 10% coupon, the 12% coupon, the 13% coupon, the 15% coupon, and the control have 9,350 samples, 29,348 samples, 2,232 samples, 333 samples, 3,414 samples, 8,840 samples, and 18,118 samples, respectively. The coupon data counts 7,681 purchasers with a revenue sum of 10,949.15€ who received a treatment and 2,435 purchasers with a revenue sum of 3,456.28€. For the Hillstrom data, we create dummy variables, which we use to replace the few factor variables to facilitate the model's predictions. Also, we clean some cryptic text strings of the customer history segment feature. No variable has solely constant or missing values. The distribution of treatments is more balanced than in the couponing case. Hence, 21,307 and 21,387 customers received the men and women campaign, respectively, while 21,306 customers did not obtain an e-mail. The Hillstrom data refers to 456 purchasers who received a treatment with a spend sum of 53,349.80€ and 122 purchasers with an aggregated revenue of 13,908.33€ from the control group. We split the data sets into five partitions of equal size and further randomly sample each partition into a 70% training and a 30% testing fold. To increase the robustness of our results, we report mean values across the data partitions.

**Results**

In the following, we exhibit the results from the causal forest's continuous response estimations and evaluation per data set and treatment. To better understand the output of the model estimations, we analyze the predicted ITE of each treatment. Figure 3 illustrates the estimated ITE distributions.

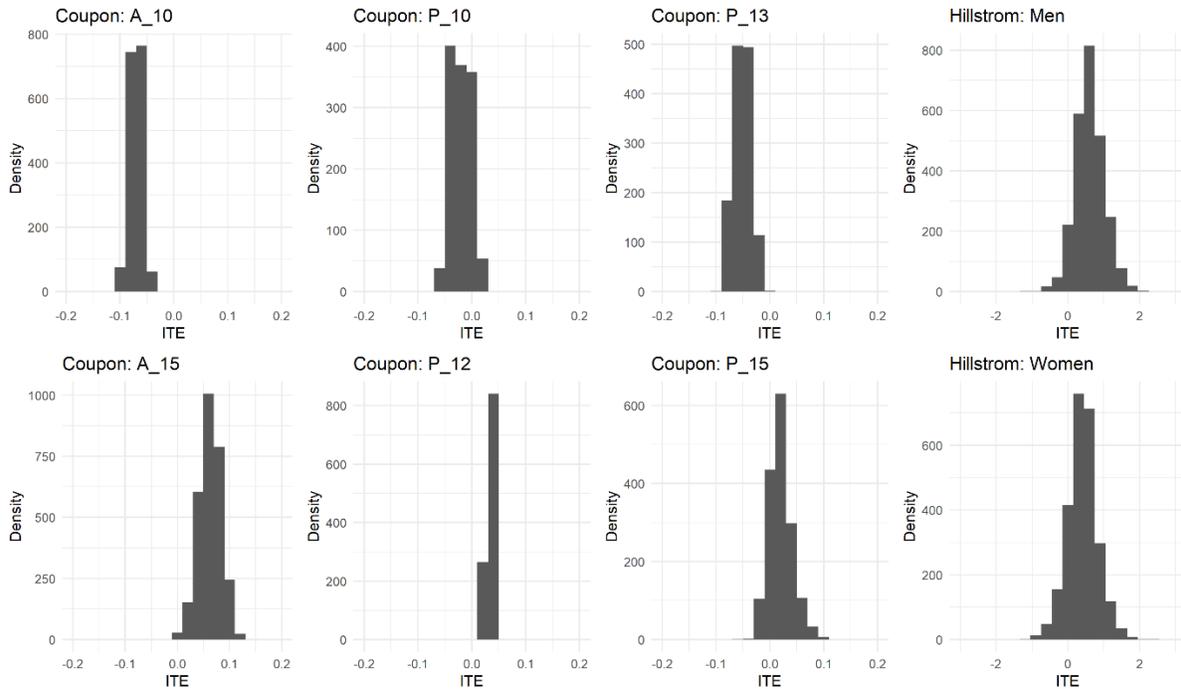

**Figure 3. Estimated ITE Distributions Per Treatment**

Based on these results, we make two observations. First, the scales of the forecasted distributions vary per data set. While the ITE distributions of the coupon data range between [-0.2; 0.2], the ITE distributions of the Hillstrom data lie in the range of [-3; 3], which suggests that the mailings have much higher treatment effects than the coupons. This is a counterintuitive finding since we would assume that customers value a coupon's financial discount higher than an advertisement without a price reduction. Second, we learn that the ITE distributions differ per coupon type. More precisely, while the coupons with a value of 10€, 10%, and 13% have mainly negative estimated ITE, the coupons with a value of 15€, 12%, and 15% have mainly positive ITE. In terms of the coupon data, the 15€ coupon has the highest ITE across the data partitions with a median of 0.07. The forecasted ITE distributions of the men and women treatments from the Hillstrom data do not significantly differ. The high (positive) ITE of both mailing types implies that many customers found the advertisements appealing and spend high purchase volumes. We remark that the women campaign has somewhat higher predicted ITE than the men campaign.

To increase the interpretability of the causal forest's predictions, we study the importance of the independent variables that were used for estimation. To increase the comparability of the most decisive variables across data sets and treatments, we apply min/max scaling. More specifically, we assign the variable with the highest importance score per treatment a value of "1" and shift the values of the remaining variables into a zero-one value range. Figure 4 displays the results of the variable importance analysis for the five most influential variables per treatment for both the coupon and Hillstrom data.

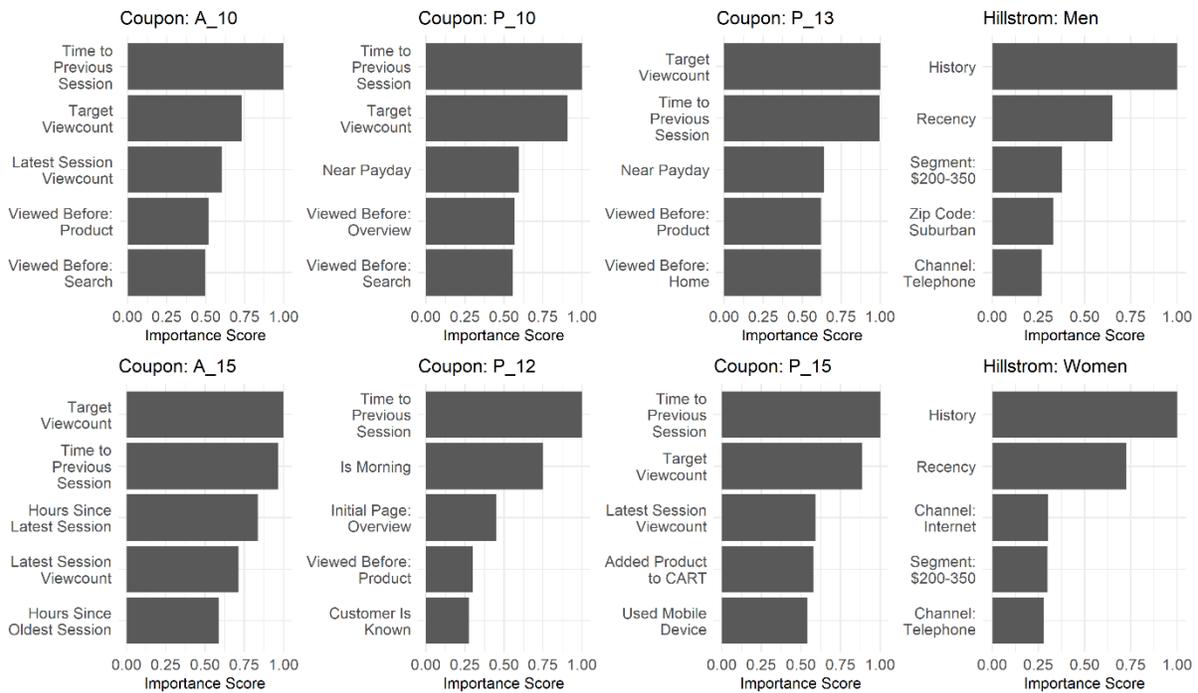

**Figure 4. Top Five Variables Per Treatment**

We derive the following findings from the variable importance analysis. Regarding the coupon data, the two most relevant variables across treatments measure the customer's period between the current and most recent session ("Time to Previous Session") and the targeted frequency of impressions from the current session ("Target Viewcount"). The 12% coupon type is the only exception in that its second most relevant variable relates to whether a visit has been made during morning times. Apart from this, some variables occur several times across treatments. These refer to the customer's previous pageviews on different page types of the shop (i.e., product, search, overview, home) and the number of impressions in the latest session. Further, some variables are critical for specific treatments. Examples include the number of hours since the latest and oldest session (from the 15€ coupon) and whether the customer is known (from the 12% coupon).

In terms of the Hillstrom data, the variables with the highest importance scores of both treatments refer to the spend amount in the previous year ("History") and the number of months since the last product purchase ("Recency"). We stress the importance of the customer's channel, segment, and location. For the men mailing, we see that important variables refer to the segment of $200-350, the affiliation to a suburban area, and the preference of telephone communications with the company. Next to the segment and telephone communication variables as for the men mailing, customer interactions through the internet receive the third rank in the context

of the women mailing. We note that the history and recency variables have much higher importance scores than the remaining variables of both mailing types.

To better understand the causal forest's evaluation per data set and treatment, we create an overview of critical business outcomes. We rank the customers according to their predicted ITE and divide the table into deciles. Per decile, we regard the number of records per treatment and control ("Ctrl" below). We state the corresponding numbers of purchasers and the amount of (scaled) revenue in total and per person. That is, we divide the spend amount of the treated and untreated customers by the number of treatment and control records, respectively. Also, we derive the incremental revenue as the difference in revenue between the treated and untreated customers ("Δ Revenue" below). We multiply the incremental revenue per person by the number of records to determine the total of incremental revenue from targeting a subpopulation. For illustration, Table 1 provides an evaluation board, which displays the results obtained from the 15€ coupon type's fifth partition for the first three deciles. These deciles are most relevant in light of budget restrictions. We choose the 15€ coupon due to its higher ITE relative to the other coupon types.

| Dec. | Records | | Purchasers | | Revenue | | | | Δ Revenue | |
|---|---|---|---|---|---|---|---|---|---|---|
| | A_15 | Ctrl | A_15 | Ctrl | Sum | | Per Person | | Per Person | Sum |
| | | | | | A_15 | Ctrl | A_15 | Ctrl | | |
| 1 | 197 | 88 | 72 | 23 | 140.40 | 48.68 | 0.71 | 0.55 | 0.16 | 45.60 |
| 2 | 168 | 117 | 27 | 18 | 34.70 | 32.48 | 0.21 | 0.28 | -0.07 | -19.95 |
| 3 | 179 | 106 | 37 | 20 | 57.50 | 30.65 | 0.32 | 0.29 | 0.03 | 8.55 |

**Table 1. Evaluation Board of the 15€ Coupon**

Table 1 shows that there are more treated than untreated customers. It further discloses that the numbers of purchasers, revenues, and incremental revenues are higher from treated customers than from untreated customers. The only exception concerns the second decile, where we see negative incremental revenues per person as untreated customers generated higher revenue on average than treated customers. Although customers from the second decile have high ITE, they would spend less than those without the incentive.

The evaluation board lacks the aspect of cumulativeness. Thus, we calculate the weighted cumulative revenue per treatment and control group. We multiply the revenue per person with the records from the treatment and control groups and cumulate accordingly. We get the incremental cumulative revenue by subtracting the cumulative revenue of the control group from its treatment group equivalent and do this per data set, partition, treatment, and decile. Moreover, we calculate median values across the ten deciles for MT-Rev and add MT-Conv as a benchmark. We further examine the proposed setting's incremental cumulative revenue in terms of the top three deciles and derive their mean values. Table 2 captures the results from the analyses.

| Data Set | Treatment | Analysis MT-Rev vs. MT-Conv | | | Analysis MT-Rev for Top Deciles | | | |
|---|---|---|---|---|---|---|---|---|
| | | MT-Rev* | MT-Conv* | % Diff. | Dec. 1 | Dec. 2 | Dec. 3 | Mean |
| Coupon | A_10 | -39.20 | -42.76 | +8.3 | -5.75 | -7.76 | -18.38 | -10.63 |
| | A_15 | 141.25 | 139.27 | +1.4 | 58.96 | 72.52 | 94.49 | 75.32 |
| | P_10 | -101.82 | -128.29 | +20.6 | -21.57 | -36.74 | -54.37 | -37.56 |
| | P_12 | -118.86 | -120.03 | +1.0 | -24.01 | -43.00 | -61.36 | -42.79 |
| | P_13 | -89.94 | -98.49 | +8.7 | -12.11 | -29.67 | -39.40 | -27.06 |
| | P_15 | -64.25 | -64.68 | +0.7 | -9.52 | -22.42 | -39.09 | -23.68 |
| Hillstrom | Men | 487.66 | 537.03 | -9.2 | 71.69 | 146.94 | 270.71 | 163.11 |
| | Women | 400.11 | 394.92 | +1.3 | -108.41 | 193.30 | 371.63 | 152.17 |

\* Median values across ten deciles

**Table 2. Results from Analyses of Incremental Cumulative Revenue**

The results reveal the following findings. First, the MT-Rev setting yields higher incremental cumulative revenues than MT-Conv for all treatments except for the mailing data set's men campaign. We remind that the MT-Conv setting comprises the estimation of a 0/1-scaled response, which here is the product purchase target variable. We argue that the financial advantage to apply the causal forest is higher if it forecasts continuous outcomes compared to binary outcomes as it better identifies likely responders to treatment with considerable spend volumes. Second, for the coupon data set, the only treatment with positive financial returns is the 15€ coupon, which is in line with prior results. In contrast, the other coupon types imply negative incremental cumulative revenue. Although the 12% coupon type has positive ITE, the treatment is not as effective as the control in terms of the incremental cumulative revenue. The same applies to the 15% treatment, which is the most effective coupon type with percentage discounts. We observe that the absolute-valued coupon types yield higher returns than their percentage-based analogs. Regarding the Hillstrom data set, we observe positive values except for the first decile of the women mailing. Although it has a lower mean value, the women campaign outperforms the men campaign on the other deciles and achieves the highest relative incremental cumulative revenue in terms of the top-decile MT-Rev analysis. The results suggest targeting customers up to the third decile with the 15€ coupon (women campaign), which would amount to additional revenue volumes of 94.49€ (371.63€) on average across the five data partitions.

Due to the substantial differences in the coupon-related returns, we interpret the related findings as follows. As the online bookseller sells mainly low-priced products, customers might perceive absolute discounts as more useful compared to percentage discounts. For instance, if a customer gets a 15€ discount off a 50€ shopping basket, s/he would need to pay 35€. In contrast, a 15% coupon would imply a final checkout payment of 42.50€, which equates to a price reduction that is half as high as for the 15€ discount type. Moreover, customers might be more comfortable with calculating the checkout prices based on absolute discounts compared to percentage discounts. It is even more effortful to calculate checkout prices based on relative coupon discounts with odd numbers (e.g., 13%) and for products with non-rounded numbers (e.g., 18.99€). Last, to explain the difference in the effectiveness of the 15€ and 10€ coupon types, customers might prefer higher absolute coupon values due to the higher price reductions.

## Conclusion, Limitations and Future Research

The paper developed revenue uplift models for multiple treatments. We argued in favor of related models as they better reflect campaign targeting practices compared to alternative models that forecast dichotomous outcomes. In light of the high complexity of non-linear algorithms, the data set requirements, and the literature's focus to develop accurate ITE estimators, our study further adds to the literature on the interpretability of an uplift model's estimation and evaluation procedures. We employed two marketing data sets with experimental data based on online coupons and mailing advertisements. Our results showed that the coupon types with absolute and percentage price reductions varied in both the stimulating effect and top-line value. The 15€ coupon type is the sole treatment with positive incremental cumulative revenue and should receive particular attention in the studied context. While the ITE distributions of the men and women campaigns did not differ much, the women campaign is somewhat of higher commercial interest based on the higher incremental cumulative revenue on the third decile. Across the ten deciles, we empirically validated the commercial advantage of MT-Rev compared to MT-Conv. As a result of the variable importance analysis, the most relevant variables per data set do not significantly differ across treatments. In particular, they measured the time to the customer's previous session and the number of view counts (coupon data) as well as the customer's spend in the recent year and the recency of the last purchase transaction (Hillstrom data).

We acknowledge the following limitations. First, we assume the online shop to apply further coupons from other channels (e.g., coupon search engines or newsletters). As only one coupon can be redeemed per session, a customer might prefer to use an alternative coupon, which implies that a competing coupon might have a larger ITE than any of the randomly allocated coupons. Given that our industry partner acts in an advisory role to the shop in the scope of the proposed coupon allocation, the data of competing coupons is unavailable. We also cannot exclude the chance that other discounts are applied in a few cases (e.g., free shipping based on club membership affiliations or gift vouchers). As this is neither displayed in the data nor is our industry partner equipped with it, we need to acknowledge this assumption as a limitation.

Future research might be interested in analyzing the effectiveness of multiple treatments for a shop's different product groups. Merchants could adapt their marketing and sales strategies accordingly to increase profits per product group. Further, an analysis of the causal forest's level of competitiveness compared to other multiple treatment revenue uplift models would be exciting.